\documentclass[conference]{IEEEtran}
\IEEEoverridecommandlockouts
\usepackage{cite}
\usepackage{amsmath,amssymb,amsfonts}
\usepackage{graphicx}
\usepackage[nolist]{acronym}
\usepackage{breqn}
\usepackage{textcomp}
\usepackage{afterpage}
\usepackage{url}
\usepackage{csquotes}
\usepackage[table]{xcolor}
\usepackage{caption}
\usepackage{subcaption}

\usepackage{algorithm}
\usepackage{algpseudocode}
\usepackage{mathtools}
\usepackage{fancyhdr}
\DeclarePairedDelimiter\ceil{\lceil}{\rceil}

\newcommand{\eqrefpref}[1]{Eq.~\eqref{#1}}

\def\BibTeX{{\rm B\kern-.05em{\sc i\kern-.025em b}\kern-.08em
    T\kern-.1667em\lower.7ex\hbox{E}\kern-.125emX}}
    
\begin{document}
\setlength{\belowcaptionskip}{-6pt}

\title{ FedAvgen: Metadata for Model Aggregation In Communication Systems \\
}

\author{
	\IEEEauthorblockN{Anthony~Kiggundu\IEEEauthorrefmark{1},~Dennis~Krummacker\IEEEauthorrefmark{1}
    ~and~Hans~D.~Schotten\IEEEauthorrefmark{1}\IEEEauthorrefmark{2}}
	\IEEEauthorblockA{
		\IEEEauthorrefmark{1}German Research Center for Artificial Intelligence (DFKI), Germany\\
		\IEEEauthorrefmark{2}RPTU University of Kaiserslautern-Landau, Germany\\
	}
}

\maketitle

\begin{abstract}
To improve business efficiency and minimize costs, \ac{AI} practitioners have adopted a shift from formulating models from scratch towards sharing pretrained models. The pretrained models are then aggregated into a global model with higher generalization capabilities, which is afterwards distributed to the client devices. This approach is known as federated learning and inherently utilizes different techniques to select the candidate client models averaged to obtain the global model. This approach, in the case of communication systems, faces challenges arising from the existential diversity in device profiles. The multiplicity in profiles motivates our conceptual assessment of a metaheuristic algorithm (FedAvgen), which relates each pretrained model with its weight space as metadata, to a phenotype and genotype, respectively. This parent-child genetic evolution characterizes the global averaging step in federated learning. We then compare the results of our approach to two widely adopted baseline federated learning algorithms like \ac{FedAvg} and \ac{FedSGD}.
\end{abstract}


\newacro{5g}[5G]{Fifth Generation}
\newacro{6g}[6G]{Sixth Generation}
\newacro{3gpp}[3GPP]{Third Generation Partnership Project}
\newacro{ue}[UE]{User Equipment}
\newacro{as}[AS]{Authentication Server}
\newacro{atm}[ATM]{Automated Teller Machine}
\newacro{ai}[AI]{Artificial Intelligence}
\newacroplural{ais}[AIs]{Artificial Intelligences}
\newacro{amm}[AMM]{Accuracy Monitoring Module}
\newacro{ga}[GA]{Genetic Algorithm}
\newacro{anlf}[AnLF]{Analytical Logical Function}
\newacro{mtlf}[MTLF]{Model Training Logical Function}
\newacro{nwdaf}[NWDAF]{Network Data Analytics Function}
\newacro{b5g}[B5G]{Beyond 5G}
\newacro{ban}[BAN]{Body Area Network}
\newacro{bsi}[\textit{BSI}]{\textit{Federal Office for Information Security}}
\newacro{bdr}[BDR]{Bit Disagreement Rate}
\newacro{bs}[BS]{Base Station}
\newacro{ca}[CA]{Certification Authority}
\newacro{cav}[CAV]{Connected Autonomous Vehicles}
\newacro{cc}[CC]{Common Criteria}
\newacro{cir}[CIR]{Channel Impulse Response}
\newacro{cr}[CR]{Challenge-Response}
\newacro{cpu}[CPU]{Central Processing Unit}
\newacro{cpps}[CPPS]{Cyber-Physical Production System}
\newacro{crl}[CRL]{Certificate Revocation List}
\newacro{csi}[CSI]{Channel State Information}
\newacro{crke}[CRKE]{Channel-Reciprocity Based Key Extraction}
\newacro{ctf}[CTF]{Channel Transfer Function}
\newacro{cotf}[COTF]{Commercial-off-the-Shelf}
\newacro{cmos}[CMOS]{Complementary Metal-Oxide-Semiconductors}
\newacro{dos}[DoS]{Denial-of-Service}
\newacro{ddos}[DDoS]{Distributed-Denial-of-Service}
\newacro{dna}[DNA]{Deoxyribonucleic Acid}
\newacro{dtls}[DTLS]{Datagram Transport Layer Security}
\newacro{dct}[DCT]{Discrete Cosine Transformation}
\newacro{ummimo}[UM-MIMO]{Ultra-Massive MIMO}
\newacro{dikw}[DIKW]{Data, Information Knowledge Wisdom}
\newacro{eal}[EAL]{Evaluation Assurance Level}
\newacro{ecc}[ECC]{Elliptic Curve Cryptography}
\newacro{ecg}[ECG]{Electrocardiogram}
\newacro{eeg}[EEG]{Electroencephalogram}
\newacro{embb}[eMBB]{enhanced Mobile broad-Band}
\newacro{emg}[EMG]{Electromyogram}
\newacro{eog}[EOG]{Electrooculography}
\newacro{enb}[eNodeB]{Evolved Node B}
\newacro{er}[ER]{Extended Reality}
\newacro{fpga}[FPGA]{Field Programmable Gate Array}
\newacro{fdd}[FDD]{Frequency Division Duplexing}
\newacro{fl}[FL]{Federated Learning}
\newacro{fedavg}[FedAvg]{Federated Averaging}
\newacro{fedsgd}[FedSGD]{Federated Stochastic Gradient Descent}
\newacro{fedavgen}[FedAvgen]{Federated Averaging Genetic}
\newacro{gdpr}[GDPR]{General Data Protection Regulation}
\newacro{gd}[G\&D]{Giesecke \& Devrient}
\newacro{h2m}[H2M]{Human-to-Machine}
\newacro{h2s}[H2S]{Human-to-Service}
\newacro{hmac}[HMAC]{Keyed-Hash Message Authentication Code}
\newacro{htc}[HTC]{Hologaphic-Type Communication}
\newacro{hotp}[HOTP]{HMAC-based One-time Password Algorithm}
\newacro{hsm}[HSM]{Hardware Security Module}
\newacro{he}[HE]{Homomorphic Encryption}
\newacro{ics}[ICS]{Industrial Control System}
\newacro{iacs}[IACS]{Industrial Automation and Control System}
\newacro{ioe}[IoE]{Internet of Everything}
\newacro{iiot}[IIoT]{Industrial Internet of Things}
\newacro{iot}[IoT]{Internet of Things}
\newacro{io}[I/O]{Input/Output}
\newacro{ic}[IC]{Integrated Circuit}
\newacro{id}[ID]{Identificator}
\newacro{ids}[IDS]{Intursion Detection System}
\newacro{irs}[IRS]{Intelligent Reflecting Surface}
\newacro{istn}[ISTN]{Integrated Space and Terrestrial Network}
\newacro{it}[IT]{Information Technology}
\newacro{itu}[ITU]{International Telecommunication Union}
\newacro{jcop}[JCOP]{Java Card Open Platform}
\newacro{kba}[KBA]{Knowledge Based Authentication}
\newacro{kdf}[KDF]{Key Derivation Function}
\newacro{led}[LED]{Light Emitting  Diode}
\newacro{lte}[LTE]{Long Term Evolution}
\newacro{ltea}[LTE-A]{Long Term Evolution Advanced}
\newacro{lr}[LR]{Linear Regression}
\newacro{los}[LoS]{Line of Sight}
\newacro{lorawan}[LoRaWAN]{Long Range Wide Area Network}
\newacro{mbb}[MBB]{Mobile Broadband}
\newacro{mfa}[MFA]{Multi-Factor Authentication}
\newacro{mcc}[MCC]{Mobile Cloud Computing}
\newacroplural{mcus}[MCUs]{Microcontroler Units}
\newacro{m2m}[M2M]{Machine-to-Machine}
\newacro{m2s}[M2S]{Machine-to-Service}
\newacro{mimo}[MIMO]{Multiple Input Multiple Output}
\newacro{mmimo}[mMIMO]{massive Multiple Input Multiple Output}
\newacro{ml}[ML]{Machine Learning}
\newacro{mulc}[mULC]{massive Ultra-Reliable Low-Latency Communication}
\newacro{mmtc}[MMTC]{massive Machine Type Communication}
\newacro{mmg}[MMG]{Mechanomyogram}
\newacro{multos}[MULTOS]{Multii-Application Smart Card Operating System}
\newacro{mux}[MUX]{Multiplexer}
\newacro{mnc}[MNC]{Mobile Network Code}
\newacro{me}[ME]{Mobile Environment}
\newacro{mac}[MACs]{Message Authentication Codes}
\newacro{mnist}[MNIST]{Modified National Institute of Standards and Technology dataset}
\newacro{ngmn}[NGMN]{Next Generation Mobile Network}
\newacro{nic}[NIC]{Network Interface Controller}
\newacro{nist}[NIST]{National Institute of Standards and Technology}
\newacro{nwdaf}[NWDAF]{Network Data Analytics Function}
\newacro{oath}[OATH]{Open Authentication}
\newacro{ocra}[OCRA]{\ac{oath} Challenge-Response Algorithm}
\newacro{ocsp}[OCSP]{Online Certificate Status Protocol}
\newacro{otp}[OTP]{One-Time Password}
\newacro{pap}[PAP]{Password-Authentication-Protocol}
\newacro{physec}[PhySec]{Physical Layer Security}
\newacro{pfs}[PFS]{Perfect Forward Secrecy}
\newacro{pin}[PIN]{Personal Identification Number}
\newacro{pkc}[PKC]{Public Key Cryptography}
\newacro{pki}[PKI]{Public Key Infrastructure}
\newacro{ppg}[PPG]{Photoplethysmography}
\newacro{prng}[PRNG]{Pseudo Random Number Generator}
\newacro{puf}[PUF]{Physically Unclonable Function}
\newacroplural{pufs}[PUFs]{Physically Unclonable Functions}
\newacro{pla}[PLA]{Physical Layer Authentication}
\newacro{psi}[PSI]{Population Stability Index}
\newacro{qr}[QR]{Quick Response}
\newacro{qos}[QoS]{Quality of Service}
\newacro{rat}[RAT]{Radio Access Technology}
\newacro{radius}[RADIUS]{Remote Authentication Dial-In User Service}
\newacro{ram}[RAM]{Random-Access Memory}
\newacro{ran}[RAN]{Radio Access Networks}
\newacro{rf}[RF]{Radio-Frequency}
\newacro{rfid}[RFID]{Radio-Frequency Identification}
\newacro{ris}[RIS]{Reconfigurable Intelligent Surface}
\newacro{rng}[RNG]{Random Number Generator}
\newacro{ro}[RO]{Ring-Oscillator}
\newacro{rom}[ROM]{Read-Only Memory}
\newacro{rs}[RS]{Reed-Solomon}
\newacro{rsa}[RSA]{Rivest-Shamir-Adleman}
\newacro{rssi}[RSSI]{Received Signal Strength Indicator}
\newacro{rsrp}[RSRP]{Reference Signal Received Power}
\newacro{re}[RE]{Resource Elements}

\newacro{sdn}[SDN]{Software-Defined Network}
\newacro{sdr}[SDR]{Software-Defined Radio}
\newacro{seccos}[SECCOS]{Secure Chip Card Operating System}
\newacro{sip}[SIP]{Session Initiation Protocol}
\newacro{skg}[SKG]{Secret Key Generation}
\newacro{sram}[SRAM]{Static Random Access Memory}
\newacro{srs}[SRS]{Software Radio Systems}
\newacro{starcos}[STARCOS]{Smart Card Chip Operating System}
\newacro{sha}[SHA]{Secure Hash Algorithm}
\newacro{se}[SE]{Static Environment}
\newacro{svm}[SVM]{Support Vector Machine}
\newacro{tcg}[TCG]{Trusted Computing Group}
\newacro{tpm}[TPM]{Trusted Platform Module}
\newacro{tls}[TLS]{Transport Layer Security}
\newacro{trng}[TRNG]{True Random Number Generator}
\newacro{tsn}[TSN]{Time-Sensitve Networking}
\newacro{tofu}[TOFU]{Trust On First Use}
\newacro{tufu}[TUFU]{Trust Upon First Use}
\newacro{totp}[TOTP]{Time-based One-time Password Algorithm}
\newacro{uav}[UAV]{Unmanned Arial Vehicles}
\newacro{usb}[USB]{Universal Serial Bus}
\newacro{usrp}[USRP]{Universal Software Radio Peripheral}
\newacro{uhd}[UHD]{USRP Hardware Driver}
\newacro{usim}[USIM]{Universal Subscriber Identity Module}
\newacro{ue}[UE]{User Equipment}
\newacro{urllc}[URLLC]{Ultra-Reliable Low-Latency Communication}
\newacro{ulbc}[ULBC]{Ultra-Reliable Low-Latency Broadband Communication}
\newacro{umbb}[uMBB]{ubiquious Mobile Broadband}
\newacro{ummimo}[UM-MIMO]{Ultra-Massive MIMO}
\newacro{vlc}[VLC]{Visible Light Communication}
\newacro{warp}[WARP]{Wireless open-Access Research Platform}

\section{Introduction}
In 5G, the \ac{3gpp} \textit{release V18.0.0} suggests that pretrained models could be shared between the analytics function (NWDAF) of one vendor's network core to another. Such analytical processes within \ac{5g} network cores of different mobile network operators \cite{3gpp.TR_23700_81} will be abstracted in the \ac{nwdaf}. This function is constituted by the \ac{mtlf}, which trains the model on a network dataset and the \ac{anlf},  which takes the trained model and evaluates it on live network input.  These virtual network functions interactively feed output into one another in individual vendor network cores which is fundamental for vendor-to-vendor aggregated model sharing and accuracy monitoring. However, as depicted by Figure \ref{fig:mltf_anlf}, pretrained models will also be shared from \ac{ue}, \ac{IoT} devices, \ac{V2X} etcetera and aggregated to get a new model with higher precision in predicting. 

\subsection{Federated Learning}

Federated Learning is a decentralized \ac{ml} approach, where models are trained across different devices using local data samples, so that the training data stays exclusively on the distributed devices. Through sharing only the model updates and an aggregation process, a global model is compiled. Different architectures for the steps of model update communication and aggregation are possible, though in the work at hand, we are covering an aggregation on a central server: Individual models are trained on multiple devices, which send the model updates to one \enquote{server}. The server performs an aggregation algorithm to compile the global model. The global model is finally transferred to the distributed clients, replacing their local models. This is usually an iterative process.
An aggregation algorithm in use can have a big influence on how well single local models are represented in the resulting global model, its performance and resource consumption of the aggregation process.
\afterpage{
  \begin{figure}
    \centering
    \includegraphics[width=0.50\textwidth]{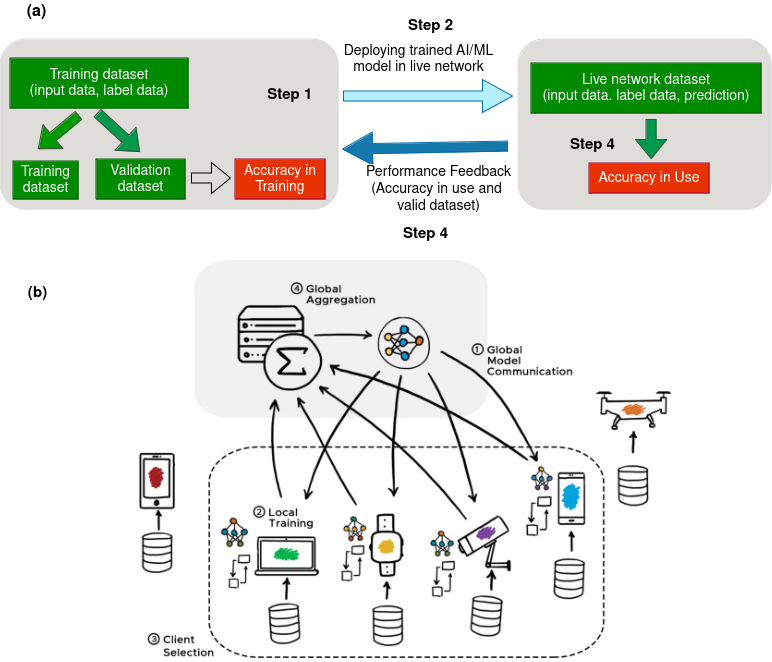}
    \caption{\small After training the model on an pre-existent dataset (in the MTLF), the model is validated using live network data in the NWDAF-AnLF component. (a) would be such a use case from one vendor network core to another. However, in (b), the device and application landscape changes such that the models shared from any device differ in feature set composition, the pre-trained model weights, and infrastructure configurations to introduce unexpected diversity. \protect\cite{Abdulwareth,3gpp.TR_23700_81}}
    \label{fig:mltf_anlf}
  \end{figure}
}%
The selection of client model updates during this global averaging process is still contentious given the heterogeneity in application and device profiles. Moreover, client updates with the global model can take place either synchronously or asynchronously \cite{you2022triple, liu2022fed2a}. Existing techniques to mitigate the effects of the resultant distortions explore adding regularization terms to client updates \cite{sannara2021federated}. 
Asynchronous aggregation has received the wider coverage given its practicality to heterogeneity \cite{sun2022asynchronous} with the discourse directed towards the selection of the client updates that characterize the aggregation procedure. 

\subsection{Goal}
While the model diversity is beneficial for the overall generalization capabilities of the models, we argue that it also complicates the aggregation of the models and inadvertently influences model inference. To catch this diversity, metadata that is descriptive of each shared model will be appended to it. Such metadata could include parameters like accuracy or loss, weights plus biases, etc.
It is still not clear however, whether as posited by \ac{3gpp}, this metadata about shared models could be the basis for model aggregation. Therefore, the applicability of known approaches (see Sec.~\ref{sec:RelWork}) for aggregation of different pre-trained models on model metadata has not been evaluated before. In addition, according to our extensive literature review, evolutionary methods have not yet been tried for the aggregation process.
The contributions of this manuscript are  therefore two-fold:

\begin{itemize}
    \item From our exhaustive literature search, evolutionary algorithms have not been explored as an alternative technique in the selection of models that form the aggregates of the global averaging process.  By exploring the model weight space, \ac{fedavgen} exploits elitism by taking advantage of the diversity in the phenotypes and genotypic formation to constitute the next generation of parents. 
    \item We assess the performance of our metaheuristic approach against canonical federated learning techniques like \ac{fedavg}\cite{mcmahan2017communication} and \ac{fedsgd} \cite{chen2016revisiting,nadiradze2021asynchronous}. In comparison to FedAvg and FedSGD, our results show that embedding the \ac{ga} (FedAvgen) in the aggregation step yields slight improvement in precision measures yet less computationally intensive. 
\end{itemize} 

We highlight in the related work section the state-of-the-art that discusses the different techniques for selecting candidate models in federated learning. This is followed by the basic details about the internals and the general composition of the \ac{ga} in the algorithm design section. We then discuss the results from our simulation runs and conclude the manuscript with a discussion about ideas for improvement of this approach.

\section{Related Work}
\label{sec:RelWork}
Generally, clustering techniques are more prevalent in recent studies, where models are aggregated based on whether they are characterized by similar statistical properties of the data distributions \cite{yan2021federated_145,bao2023optimizing_147}. These similarities have also been premised on geospatial relationships defined in terms of distance measures like the Manhattan, Euclidean or cosine distance \cite{palihawadana2022fedsim_140}. These kind of geographic divisions yield local clusters of models \cite{9771904_144} that are bound by communication capabilities to form hierarchies of partial aggregations that later comprise the global averaging \cite{9562522_139,9918588_146}.
Temporal dynamic fusion studies \cite{chen2022dynamic} propose periodic windowing or restricting the number of participants in each aggregation round \cite{lee2021adaptive, liu2021fedpa}. This periodic selection has been defined around adaptive deadlines \cite{zhang2021dynamic, huba2022papaya} such that updates submitted beyond a preset time threshold are considered stale and excluded in the aggregation. Other such temporal schemes take into account the order in which updates arrive in a particular computation round \cite{9562538} or prioritizing some models over others using a prior calculated index \cite{9359185}. \cite{ wu2021fast}'s experiments involved adaptively assigning different weights to client updates during the aggregation.

Infrastructure based performance-oriented metrics have also been employed for selective aggregation procedures  \cite{9810502} while other findings evaluate this candidacy based on client reputations (probability of task completion for example) \cite{wang2019adaptive, wang2021reputation}. Clients have also been selected based on rewards attained from active participation and contribution \cite{lim2021decentralized_132,hui2022quality_133}. 
\cite{9562751}'s studies proposed optimizing the number of local models to aggregate using gradient descent routines in their online learning algorithm. Similarly, \cite{Boxin} proposes an online algorithm that learns from bandit feedback to intelligently sample clients for aggregation in a stochastic gradient descent approach.
Information gain has also been deployed in client sampling for aggregation to reduce the communication burden pre-trained model sharing introduces in the system \cite{shukla2021federated_221}. 
\cite{9593194} did some comparative performance evaluations, revealing  differences in aggregations techniques for both synchronous and asynchronous federations and argued for periodic model integration to mitigate the straggler effect. While \ac{fedsgd} like \ac{fedavg} randomly samples the population of models to select a predefined portion of these that participate in the global averaging. And recently, studies propose checkpoint merging with Bayesian optimization in large language models \cite{liu2024checkpoint}. Other findings suggest knowledge distillation approaches where sub-networks are trained to learn specific local phenomena and the sub-networks later dispatch only this knowledge to the global averaging server \cite{he2020group}.

\section{Algorithm Design} 
\subsection{Genetic Algorithms}
\acp{ga} are an optimization technique, inspired by natural selection and genetics as known from biology. By mimicking the evolutionary process, they aim to find approximate solutions, as a set of known specimen (potential solutions; the \textit{initial population}) is used to combine partial properties of individuals to generate new specimen -- resulting in a new \textit{generation}.
For evolving a new generation, the \textit{parents} can be selected from previous generations. Selection of parents is usually based on a \textit{fitness score} to continue evolution with the best available properties. For quantifying this, a \textit{fitness function} is to be defined depending on the problem formulation, denoted $F(Q_i)$ ($Q_i$ being the individual of index $i$).
The term \textit{genotype} specifies how a specimen is encoded, giving its properties.
\textit{Chromosomes} are partial information packages, of which the entire genotype is comprised. The \textit{phenotype} is the expression of the genes in the problem space. With a \ac{ga}, it is derived from the genotype through a mapping process, according to $F(Q_i)$.
The production of an offspring is performed by a \textit{crossover operation}, which specifies how chromosomes are selected from parents and combined to the offspring.
It is an iterative process, with the goal to increase the results of $F(Q_i)$ throughout generations by chance. Termination criterion can be a threshold of $F(Q_i)$, number of generations, the derivative of $F(Q_i)$'s results, etc.

\subsection{Pre-trained Model Metadata}
Denoting a set  $S={Q_{1},Q_{2}, \ldots Q_{N}}$ of $N$ pre-trained models, classic propositions in federated learning define for the global objective function as follows: 
\begin{equation}
    \min\limits_{Q_{S}} F(Q_{S}) = \sum^{\tau N}_{i=1} I_{i}\delta_{i}(Q_{i})
    \label{lbl:fedavg}
\end{equation}
Here, $\tau$ determines the portion of phenotypes (models) that will parent the next generation in each epoch. This portion is weighted based on the importance $I_{i}$ of the phenotype and is usually a factor of the underlying dataset.
$\delta_{i}$ defines the objective function local to the phenotype, and encapsulates the loss in training. According to \cite{mcmahan2017communication}, \eqrefpref{eqn:global} is then definitive of the global averaged phenotype $Q_{S}$ over all candidates $N$.
\begin{equation}
   Q_{S} = \sum^{\tau N}_{i=1} I_{i}Q_{i}
    \label{eqn:global}
\end{equation}
The \textit{FedAvg} in our setup does a randomized sampling of this $\tau$ from all $N$ to ensure some diversity in the global averaging participants.
The participants in this case were the pre-trained $ MobileNet\_V3\_Small$ \cite{Howard2017MobileNetsEC} models provided at known model repository Huggingface. These shared pre-trained models ship with some performance related metadata like the model accuracy or loss, a weight and bias space, model type, learning rate, optimizers, number of epochs, local batch size etc. 

Drawing this same analogy, we can denote as $G$ this set of metadata about the pre-trained model.
In our \ac{ga} \textit{FedAvgen}, each model denotes a phenotype characterized by the metadata as genes (genotype). The structure of the phenotype $Q_{i}$ therefore comprises elements $Q_{i,m} = \{m_{1}, m_{2}, \ldots m_{G}\}$ where, $m_{1}, m_{2}$ and $m{3}$ here denote the three qualitative weight space measures sparsity, stability and health. 

\subsection{Composing the Fitness Function} 
From the literature, three metrics about the weight space exist to qualitatively evaluate how generalizable the model is to previously unseen inputs. We defined the fitness of an evolved genetic encoding around the following weight space specific measures.

\textbf{Sparsity} $\zeta$: This measure (a numeric value between $[0,1]$) indicates a fraction of weights that are near zero. For a weight vector $w$ of the entire model weight space $L_{Q_{i}}$ therefore, we use Hoyer's \cite{Hoyer} sparsity metric for this weight vector $w$. This is given its robustness and scale-invariance in continuous time to yield a smooth and normalized measure.  \eqrefpref{eqn:sparse} is the formulation for the Gini Index of the sparsity metric. 
 \begin{equation} %
     \zeta(w) = \frac{\sqrt{L_{Q_{i}}}-{\frac{||w||_{1}}{||w||_{2}}}} {\sqrt{L_{Q_{i}}-1}}
     \label{eqn:sparse}
 \end{equation}
 \textit{where $||w||_{1}$  and $||w||_{2}$ denote the sum of absolute values and the Euclidean norm. $\frac{||w||_{1}}{||w||_{2}}$ measures the density of the weights. The denominator is a norm that bounds the metric between the intervals $0$ (dense) and $1$ (very sparse with only one non-zero weight). }
     
\textbf{Weight health} $\eta$: Ideally, this metric shows how close the distribution of the weights is to a target distribution and the objective is to have weights whose distribution is closer to the target. We can ascertain this measure using the mean and standard deviation, where given $w\in \mathbb{R}_{K}$ as the weight vector of model $Q_{i}$, we compute these measures using \eqrefpref{eqn:mean_std}.
\begin{equation}
    \begin{split}
        \mu_{w} = \frac{1}{K} \sum_{i=1}^{K} w_{i} \\
        \sigma_{w} = \sqrt{\frac{1}{K}\sum_{i=1}^{K}(w_{i}-\mu_{w})^{2}}
    \end{split}
\label{eqn:mean_std}
\end{equation}
 We then normalize both the standard deviation and the mean around the target to express for the resultant weight health score using \eqrefpref{eqn:health}.
 \begin{equation}
     \eta(w) = - \bigg ( \frac{|\mu_{w}|}{\sigma_{target}} + \frac{|\sigma_{w} - \sigma_{target}|}{\sigma_{target}}\bigg)
     \label{eqn:health}
 \end{equation}
 \textit{where $|\mu_{w}|$ measures how the weights differ from $0$ and $|\sigma_{w} - \sigma_{target}|$ measures the deviation of the standard deviation from the target weights.}

\textbf{Algorithmic Stability} $\rho$: In learning theory, this metric is an indicator for how impaired the gradient is and robustness to perturbations. In essence, this enables evaluating how introducing small changes in the input space affects the target predictions. The weights are considered stable when small or big gradients in the input have almost negligible effect on the precision in target predictions. When weight vector between two subsequent epochs are analyzed, the closer to 1 the difference between these weight vectors is, the higher the stability of the weights and vice versa. Therefore, given two weight vectors $w^{t}$ and $w^{t-1}$ at epoch count $t$ and $t-1$ respectively, we compute the stability $\rho$ of the weights using \eqrefpref{eqn:stable}.
\begin{equation}
    \rho(w) = 1 - \frac{||w^{(t)} - w^{(t-1)}||_{2}} {||w^{(t-1)}||_{2}}
    \label{eqn:stable}
\end{equation}
\textit{where the denominator normalizes the Euclidean norm of the deviation between the weights vectors ($w^{(t)}$ and $w^{(t-1)}$). }  

The eligibility of each phenotype to participate in parenthood is then determined by a fitness function. This function encapsulates the above three qualitative measures ($\epsilon, \beta, \gamma$) as genotypes that characterize a phenotype $Q_{i}$, and this function defined by  \eqref{eqn:fitness} assigns a fitness score to $Q_{i}$. Depending on this score, $Q_{i}$ will participate in or be dropped from parenthood.

    \begin{equation}
            F(Q_{i}) = \epsilon(1-\zeta)+\beta\rho + \gamma\frac{1}{\eta} 
    \label{eqn:fitness}
    \end{equation}
    \textit{where $\epsilon, \beta, \gamma$ are hyper-parameters denoting importance weights with monotonically decreasing coefficients. }

We begin by loading the pre-trained models and then for each model we extract the relevant metadata as classical qualitative indicators about that model's weight space. This metadata is then used to comprise the initial population. Using uniform crossover as the tournament operator, a pair of parents are chosen in each subsequent generation.  The mutation operation of this surrogate model is characterized by random noise generated from a uniform distribution, which is added to the genomes and followed by some rescaling to maintain the Hamming weight.
Then an evaluation of the genetic sequencing of the evolving surrogate is performed by the fitness function $F(Q_{i})$. 
Because the objective is to retain models with high precision at each computation step, we the genetic algorithm here embeds elitist routines that are less exploratory. This ensures that quality models are not discarded as would be the case in some random chancing tournament procedures. We denote the population of all phenotypes at a given generation $t$ as $P^{t} = {Q_{1}^{t}, Q_{2}^{t}, \ldots , Q_{N}^{t}}$ each with fitness measure $F(Q_{i}^{t})$. Such that, \eqrefpref{eqn:reorder} characterizes the reordering of the population in decreasing order of their fitness measures using a permutation $\alpha$ on the indices. 
\begin{equation}
    F(Q_{\alpha(1)}) \geq F(Q_{\alpha(2)}) \geq  \ldots \geq F(Q_{\alpha(N)})
    \label{eqn:reorder}
\end{equation}
An arbitrarily set elite rate \{ $\lambda \in \mathbb{R}: 0 \leq \lambda \leq 1$ \} then selects the elite phenotypes $\mathbf{E} = \ceil*{\lambda N}$ to form the elite set defined by \eqrefpref{eqn:elite}.
\begin{equation}
    \mathbf{E}^{t} =  Q_{\alpha(1)}^{t}, Q_{\alpha(2)}^{t}, \ldots , Q_{\alpha(E)}^{t} 
    \label{eqn:elite}
\end{equation}
Such that, \eqrefpref{eqn:parents} characterized for the candidates $P^{t+1}$ for aggregating from the elite set $\mathbf{E}^{t}$ as:

\begin{equation}
    P^{t+1} = \sum_{i\in\mathbf{E}^{t}} \frac{F(Q_{i})} {\sum_{j\in \mathbf{E}^{t}}F(Q_{j})}Q_{i}^{(t)}
    \label{eqn:parents}
\end{equation}

\section{Results and Evaluation}
\begin{figure*} 
    \centering
            \makebox[\textwidth]{\includegraphics[width=0.577\paperwidth]{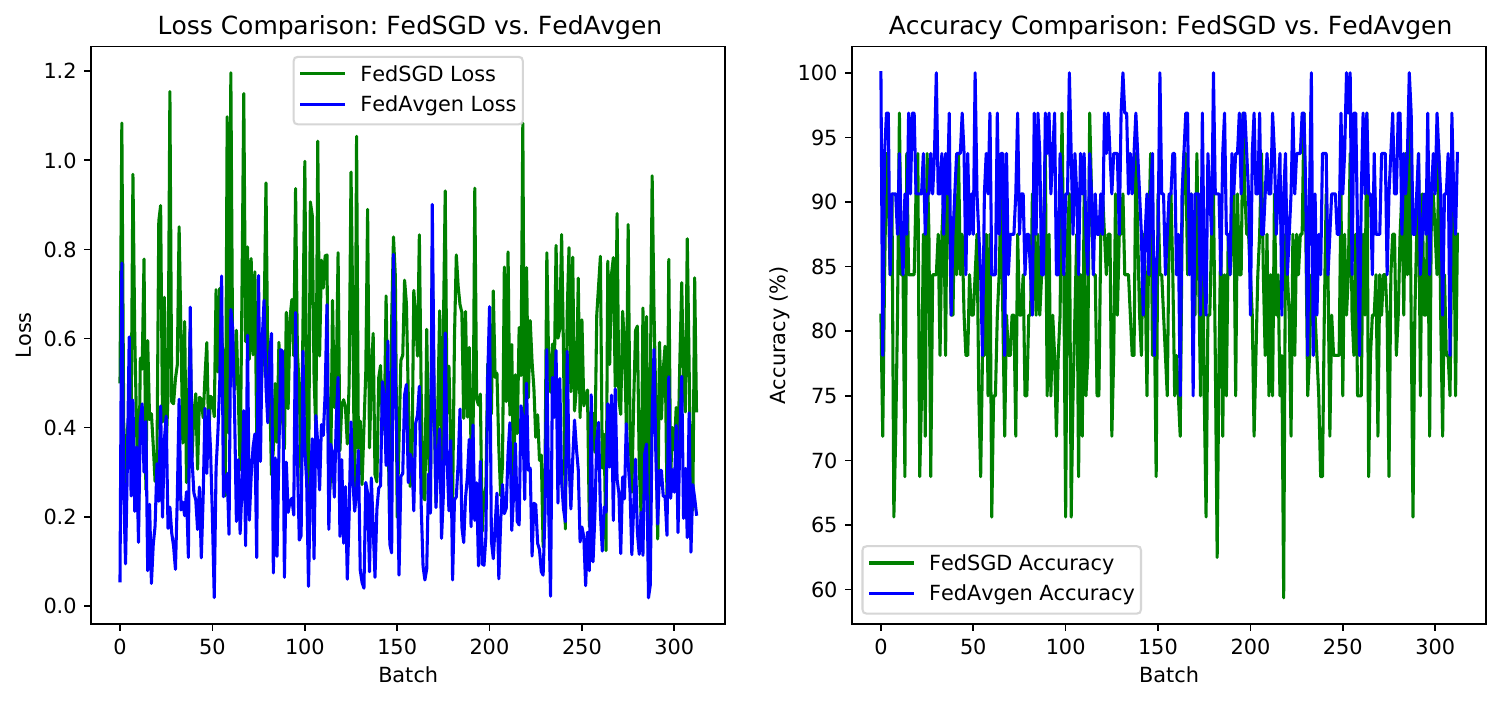}}
       \caption{\small The generalization capabilities of the FedAvgen algorithm against FedSGD with a lower final measure in the loss value.}       
    \label{fig:acc_loss}
\end{figure*}

\begin{figure*}
   \centering
   \makebox[\textwidth]{\includegraphics[width=0.575\paperwidth]{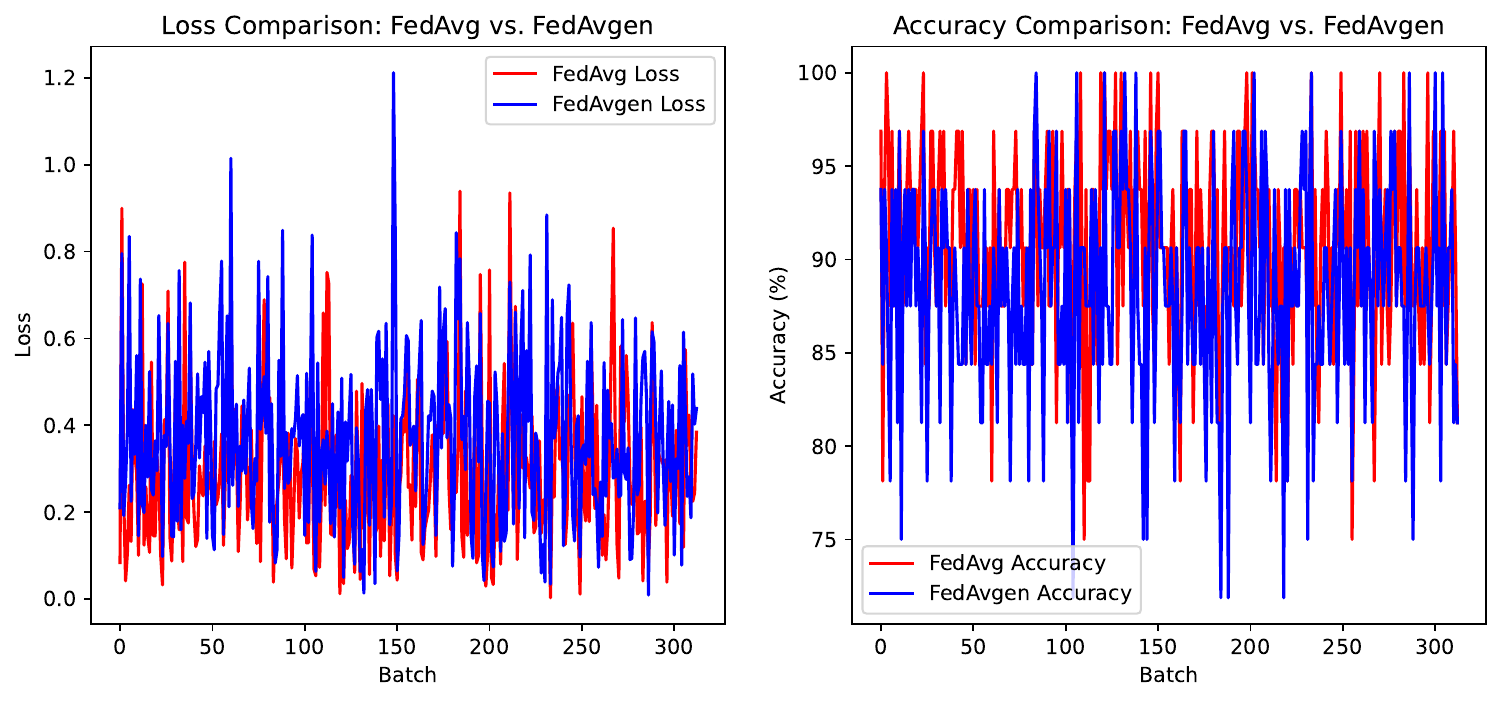}}
     \caption{\small In comparison to FedAvg, the FedAvgen algorithm gives better prediction capabilities.}
     
 \label{fig:sgd_acc_loss}
\end{figure*}

\begin{figure*} 
    \centering
       \begin{subfigure}[b]{0.465\textwidth} 
           \includegraphics[width=\textwidth]{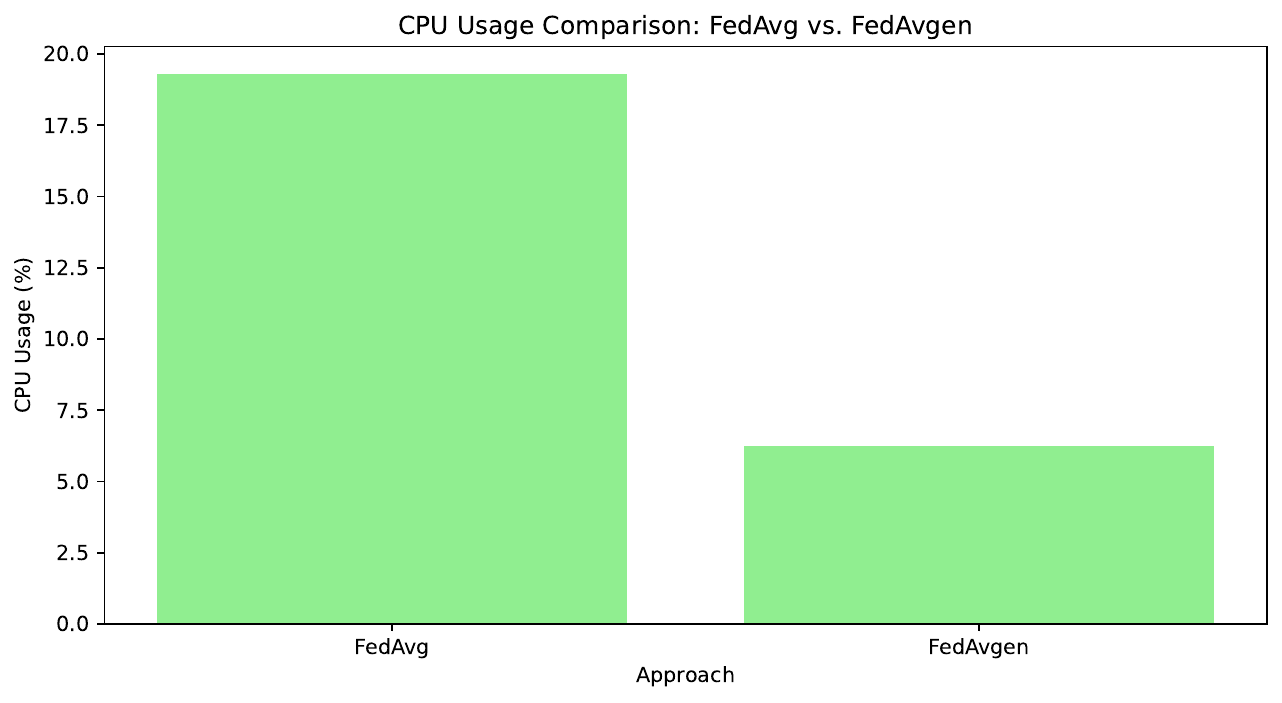}
       \end{subfigure}
        \begin{subfigure}[b]{0.465\textwidth} 
           \includegraphics[width=\textwidth]{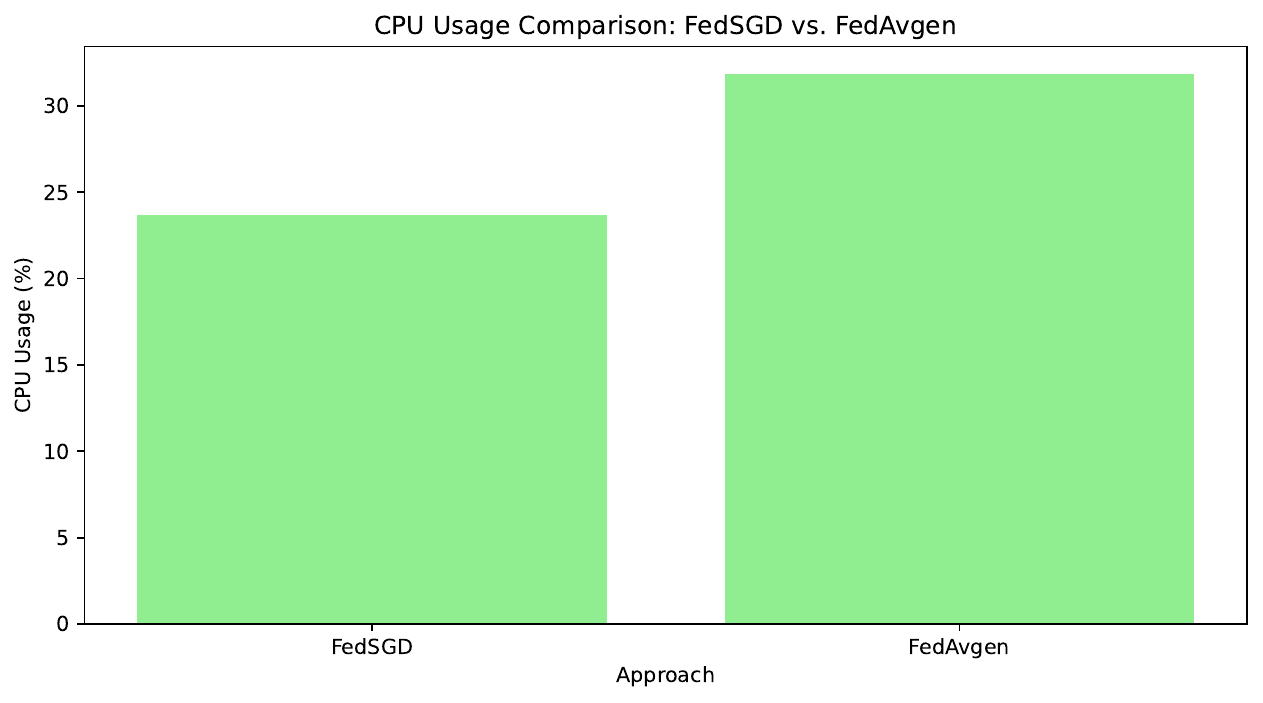}
       \end{subfigure}
       \caption{\small The $FedAvgen$ here is less computationally intensive in comparison to $FedAvg$ but requires more processor cycles than the $FedSGD$ algorithm. }
       
    \label{fig:resources}
    \end{figure*}

The globally averaged models were evaluated on the CIFAO-10 dataset \cite{Krizhevsky2009LearningML}, evaluated over 300 batches in 5 epochs. 
Table \ref{table:fedparams} summarizes the parameters used for both the \textit{FedAvg, FedSGD} while Table \ref{table:genparams} captures parameters specific to the \textit{FedAvgen} experiments. Essentially, here the $FedAvg$ samples the clients that participate in the global averaging step randomly. In Figure \ref{fig:acc_loss} and \ref{fig:sgd_acc_loss}, it is evident that, because the elitism ensures only models with the highest gradients qualify in the parenting, the $FedAvgen$ records slightly higher accuracy and converges faster with lower measures in the loss. This means $FedAvgen$ is more robust to predicting phenomena given unseen input. This can be attributed to the algorithm's adaptive client selection procedures unlike $FedAvg$ or $FedSGD$ that treats the clients equally and randomizes the selection. Discarding low accuracy updates from clients yields better gradients and minimizes oscillations leading to faster convergence.
Figure \ref{fig:resources} on the other hand depicts the fraction of the processor that the different algorithms consume at runtime. Because $FedSGD$ forwards gradients to the global servers, the clients do less computation than the $FedAvgen$ that does most computation locally to reduce the communication overhead. It could be therefore postulated that $FedSGD$ is more suited for devices limited by energy capacity.

\begin{table}[]
   \rowcolors{1}{gray!50}{white}
     \caption{ The $FedAvg$ and $FedSGD$ algorithms in our setup were deployed with the same parameters}
    \centering
    \begin{tabular}{ |p{3cm}||p{2cm}||p{2cm}|  }
    \hline
        \multicolumn{3}{|c|}{\textbf{FedAvg and FedSGD}} \\
    \hline
        \textbf{Parameter} & \textbf{FedAvg} & \textbf{FedSGD}\\
    \hline
        Learning rate   & 0.001  & 0.001  \\
        Optimizer&   Adam &   Adam \\
        Criterion & CrossEntropy & CrossEntropy\\
        Batch size    & 32 & 32 \\
        Epochs &   5 &   5\\ 
        Num. pre-trained models &   30 &   30 \\
        \hline
    \end{tabular}    
    \label{table:fedparams}
\end{table}

\begin{table}[]
   \rowcolors{1}{gray!50}{white}
     \caption{ The tabulation of the genetic algorithm specific parameters}
    \centering
    \begin{tabular}{ |p{4cm}||p{3.5cm}| }
    \hline
        \multicolumn{2}{|c|}{\textbf{FedAvgen parameter set}} \\
    \hline
        \textbf{Parameter} & \textbf{Value} \\
    \hline
        Fitness weight coeffs. $\epsilon$,$\beta$,$\gamma$ respectively   & 0.5 (sparsity),  0.3 (stability),  0.2 (health)   \\
        Genetic Operators &   uniform crossover \\
        Mutation rate & 0.01 \\
        Num. pre-trained models &   30 \\
        Num. selected models    & 2 \\
        Elite rate & 2 \\
    \hline
    \end{tabular}    
    \label{table:genparams}
\end{table}

\section{Discussion and Conclusion}
The devices that will comprise the \ac{6G} ecosystem are so diverse in their profiles. And so will the pretrained models shared for integration into a global model using federated learning techniques. The challenges associated with this heterogeneity are the drive to research questions about selecting candidate pretrained models that participate in the global averaging. This manuscript explores biologically inspired approaches for the selection of these candidates in contrast to existing techniques that either cluster or randomly select the candidates.
Our results show that it is indeed possible to use the metadata about the pretrained shared models to aggregate them while maintaining the predictive capabilities. Moreover, embedding regularization techniques like $L2$ weight decaying and normalization would reduce the overfitting and improve the generalization.
However, re-using the models becomes challenging if clear descriptive information (metadata) about the pretrained models is missing. This metadata can be collected about artifacts like models, datasets, training configurations, evaluation etc.
It is therefore inevitable that appending to the shared models qualitative statistical measures about the underlying datasets could improve the model aggregation process. Such measures could include 
metrics about the dataset drift, missing fields, size or checksum etc. These metrics should then contribute to the aggregates selection during the global averaging routines. 
Additional measures like context and concept drifts would also help tell when features in the underlying datasets are no longer relevant or appropriate for the forecasting problem at hand. 
\section*{Acknowledgment}

This work is supported in fully by the German Federal Ministry of Education and Research (BMBF) within the Open6GHub project under grant numbers \textit{16KISK003K and 16KISK004}.

\begin{acronym}[HRTEM]
  \acro{QoS}{Quality of Service}
  \acro{QoE}{Quality of Experience}
  \acro{AnLF}{Analytics logical function}
  \acro{AI}{Artificial Intelligence}
  \acro{NSSF}{Network Slice Selection Function}
  \acro{SMF}{Session Management Function}
  \acro{SDN}{Software Defined Networks}
  \acro{3GPP}{Third Generation Partnership Project}
  \acro{FCFS}{First Come First Server}
  \acro{LCFS}{Last Come First Serve}
  \acro{V2X}{Vehicle to Everything}
  \acro{SBA}{Service based Architectures}
  \acro{5G}{Fifth Generation}
  \acro{6G}{Sixth Generation}
  \acro{ONNX}{Open Neural Network Exchange}
  \acro{M/M/C}{Markovian/ Markovian/ number of queues}
  \acro{AIMMX}{Automated AI Model Metadata eXtractor}
  \acro{M/G/C}{Markovian/General/ number of queues}
  \acro{G/G/C}{General/ General/ number of queues}
  \acro{RAN}{Radio Access Network}
  \acro{IoT}{Internet of Things}
  \acro{O-RAN}{Open Radio Access Network}
  \acro{DU}{Distributed Unit}
  \acro{RU}{Radio Unit}
  \acro{CU}{Centralized Unit}
  \acro{SIRO}{Serve In Random Order}
  \acro{FedAvg}{Federated Averaging}
  \acro{FedSGD}{Federated Stochastic Gradient Descent}
  \acro{MNIST}{Modified National Institute of Standards and Technology dataset}
\end{acronym}
\bibliographystyle{IEEEtran}
\bibliography{bibfile/GA-FevAvg.bib}{}

\begin{thebibliography}{10}
\providecommand{\url}[1]{#1}
\csname url@samestyle\endcsname
\providecommand{\newblock}{\relax}
\providecommand{\bibinfo}[2]{#2}
\providecommand{\BIBentrySTDinterwordspacing}{\spaceskip=0pt\relax}
\providecommand{\BIBentryALTinterwordstretchfactor}{4}
\providecommand{\BIBentryALTinterwordspacing}{\spaceskip=\fontdimen2\font plus
\BIBentryALTinterwordstretchfactor\fontdimen3\font minus
  \fontdimen4\font\relax}
\providecommand{\BIBforeignlanguage}[2]{{%
\expandafter\ifx\csname l@#1\endcsname\relax
\typeout{** WARNING: IEEEtran.bst: No hyphenation pattern has been}%
\typeout{** loaded for the language `#1'. Using the pattern for}%
\typeout{** the default language instead.}%
\else
\language=\csname l@#1\endcsname
\fi
#2}}
\providecommand{\BIBdecl}{\relax}
\BIBdecl

\bibitem{3gpp.TR_23700_81}
3GPP, ``Technical specification group services and system aspects; study of
  enablers for network automation for the 5g system (5gs),''
  \url{https://portal.3gpp.org/desktopmodules/Specifications/SpecificationDetails.aspx?specificationId=4011},
  3rd Generation Partnership Project (3GPP), Technical Specification (TS)
  23.700-81, 12 2022, version 18.0.0.

\bibitem{you2022triple}
L.~You, S.~Liu, Y.~Chang, and C.~Yuen, ``A triple-step asynchronous federated
  learning mechanism for client activation, interaction optimization, and
  aggregation enhancement,'' \emph{IEEE Internet of Things Journal}, vol.~9,
  no.~23, pp. 24\,199--24\,211, 2022.

\bibitem{liu2022fed2a}
S.~Liu, Q.~Chen, and L.~You, ``Fed2a: Federated learning mechanism in
  asynchronous and adaptive modes,'' \emph{Electronics}, vol.~11, no.~9, p.
  1393, 2022.

\bibitem{sannara2021federated}
E.~Sannara, F.~Portet, P.~Lalanda, and V.~German, ``A federated learning
  aggregation algorithm for pervasive computing: Evaluation and comparison,''
  in \emph{2021 IEEE international conference on pervasive computing and
  communications (PerCom)}.\hskip 1em plus 0.5em minus 0.4em\relax IEEE, 2021,
  pp. 1--10.

\bibitem{sun2022asynchronous}
Y.~Sun, J.~Shao, Y.~Mao, and J.~Zhang, ``Asynchronous semi-decentralized
  federated edge learning for heterogeneous clients,'' in \emph{ICC 2022-IEEE
  International Conference on Communications}.\hskip 1em plus 0.5em minus
  0.4em\relax IEEE, 2022, pp. 5196--5201.

\bibitem{Abdulwareth}
\BIBentryALTinterwordspacing
O.~R.~A. Almanifi, C.-O. Chow, M.-L. Tham, J.~H. Chuah, and J.~Kanesan,
  ``Communication and computation efficiency in federated learning: A survey,''
  \emph{Internet of Things}, vol.~22, p. 100742, 2023. [Online]. Available:
  \url{https://www.sciencedirect.com/science/article/pii/S2542660523000653}
\BIBentrySTDinterwordspacing

\bibitem{mcmahan2017communication}
B.~McMahan, E.~Moore, D.~Ramage, S.~Hampson, and B.~A. y~Arcas,
  ``Communication-efficient learning of deep networks from decentralized
  data,'' in \emph{Artificial intelligence and statistics}.\hskip 1em plus
  0.5em minus 0.4em\relax PMLR, 2017, pp. 1273--1282.

\bibitem{chen2016revisiting}
J.~Chen, X.~Pan, R.~Monga, S.~Bengio, and R.~Jozefowicz, ``Revisiting
  distributed synchronous sgd,'' \emph{arXiv preprint arXiv:1604.00981}, 2016.

\bibitem{nadiradze2021asynchronous}
G.~Nadiradze, A.~Sabour, P.~Davies, S.~Li, and D.~Alistarh, ``Asynchronous
  decentralized sgd with quantized and local updates,'' \emph{Advances in
  Neural Information Processing Systems}, vol.~34, pp. 6829--6842, 2021.

\bibitem{yan2021federated_145}
Z.~Yan, Y.~Z. Yi, Z.~JiLin, Z.~NaiLiang, R.~YongJian, W.~Jian, and Y.~Jun,
  ``Federated learning model training method based on data features perception
  aggregation,'' in \emph{2021 IEEE 94th Vehicular Technology Conference
  (VTC2021-Fall)}.\hskip 1em plus 0.5em minus 0.4em\relax IEEE, 2021, pp. 1--7.

\bibitem{bao2023optimizing_147}
W.~Bao, H.~Wang, J.~Wu, and J.~He, ``Optimizing the collaboration structure in
  cross-silo federated learning,'' in \emph{International Conference on Machine
  Learning}.\hskip 1em plus 0.5em minus 0.4em\relax PMLR, 2023, pp. 1718--1736.

\bibitem{palihawadana2022fedsim_140}
C.~Palihawadana, N.~Wiratunga, A.~Wijekoon, and H.~Kalutarage, ``Fedsim:
  Similarity guided model aggregation for federated learning,''
  \emph{Neurocomputing}, vol. 483, pp. 432--445, 2022.

\bibitem{9771904_144}
Y.~Sun, J.~Shao, Y.~Mao, J.~H. Wang, and J.~Zhang, ``Semi-decentralized
  federated edge learning for fast convergence on non-iid data,'' in \emph{2022
  IEEE Wireless Communications and Networking Conference (WCNC)}, 2022, pp.
  1898--1903.

\bibitem{9562522_139}
F.~P.-C. Lin, S.~Hosseinalipour, S.~S. Azam, C.~G. Brinton, and N.~Michelusi,
  ``Semi-decentralized federated learning with cooperative d2d local model
  aggregations,'' \emph{IEEE Journal on Selected Areas in Communications},
  vol.~39, no.~12, pp. 3851--3869, 2021.

\bibitem{9918588_146}
Z.~Li, Z.~Chen, X.~Wei, S.~Gao, C.~Ren, and T.~Q. Quek, ``Hpfl-cn:
  Communication-efficient hierarchical personalized federated edge learning via
  complex network feature clustering,'' in \emph{2022 19th Annual IEEE
  International Conference on Sensing, Communication, and Networking (SECON)},
  2022, pp. 325--333.

\bibitem{chen2022dynamic}
Z.~Chen, H.~Cui, E.~Wu, and X.~Yu, ``Dynamic asynchronous anti poisoning
  federated deep learning with blockchain-based reputation-aware solutions,''
  \emph{Sensors}, vol.~22, no.~2, p. 684, 2022.

\bibitem{lee2021adaptive}
J.~Lee, H.~Ko, and S.~Pack, ``Adaptive deadline determination for mobile device
  selection in federated learning,'' \emph{IEEE Transactions on Vehicular
  Technology}, vol.~71, no.~3, pp. 3367--3371, 2021.

\bibitem{liu2021fedpa}
J.~Liu, J.~H. Wang, C.~Rong, Y.~Xu, T.~Yu, and J.~Wang, ``Fedpa: An adaptively
  partial model aggregation strategy in federated learning,'' \emph{Computer
  Networks}, vol. 199, p. 108468, 2021.

\bibitem{zhang2021dynamic}
W.~Zhang, T.~Zhou, Q.~Lu, X.~Wang, C.~Zhu, H.~Sun, Z.~Wang, S.~K. Lo, and F.-Y.
  Wang, ``Dynamic-fusion-based federated learning for covid-19 detection,''
  \emph{IEEE Internet of Things Journal}, vol.~8, no.~21, pp. 15\,884--15\,891,
  2021.

\bibitem{huba2022papaya}
D.~Huba, J.~Nguyen, K.~Malik, R.~Zhu, M.~Rabbat, A.~Yousefpour, C.-J. Wu,
  H.~Zhan, P.~Ustinov, H.~Srinivas \emph{et~al.}, ``Papaya: Practical, private,
  and scalable federated learning,'' \emph{Proceedings of Machine Learning and
  Systems}, vol.~4, pp. 814--832, 2022.

\bibitem{9562538}
Q.~Ma, Y.~Xu, H.~Xu, Z.~Jiang, L.~Huang, and H.~Huang, ``Fedsa: A
  semi-asynchronous federated learning mechanism in heterogeneous edge
  computing,'' \emph{IEEE Journal on Selected Areas in Communications},
  vol.~39, no.~12, pp. 3654--3672, 2021.

\bibitem{9359185}
J.~Hao, Y.~Zhao, and J.~Zhang, ``Time efficient federated learning with
  semi-asynchronous communication,'' in \emph{2020 IEEE 26th International
  Conference on Parallel and Distributed Systems (ICPADS)}, 2020, pp. 156--163.

\bibitem{wu2021fast}
H.~Wu and P.~Wang, ``Fast-convergent federated learning with adaptive
  weighting,'' \emph{IEEE Transactions on Cognitive Communications and
  Networking}, vol.~7, no.~4, pp. 1078--1088, 2021.

\bibitem{9810502}
A.~Sultana, M.~M. Haque, L.~Chen, F.~Xu, and X.~Yuan, ``Eiffel: Efficient and
  fair scheduling in adaptive federated learning,'' \emph{IEEE Transactions on
  Parallel and Distributed Systems}, vol.~33, no.~12, pp. 4282--4294, 2022.

\bibitem{wang2019adaptive}
S.~Wang, T.~Tuor, T.~Salonidis, K.~K. Leung, C.~Makaya, T.~He, and K.~Chan,
  ``Adaptive federated learning in resource constrained edge computing
  systems,'' \emph{IEEE journal on selected areas in communications}, vol.~37,
  no.~6, pp. 1205--1221, 2019.

\bibitem{wang2021reputation}
Y.~Wang and B.~Kantarci, ``Reputation-enabled federated learning model
  aggregation in mobile platforms,'' in \emph{ICC 2021-IEEE international
  conference on communications}.\hskip 1em plus 0.5em minus 0.4em\relax IEEE,
  2021, pp. 1--6.

\bibitem{lim2021decentralized_132}
W.~Y.~B. Lim, J.~S. Ng, Z.~Xiong, J.~Jin, Y.~Zhang, D.~Niyato, C.~Leung, and
  C.~Miao, ``Decentralized edge intelligence: A dynamic resource allocation
  framework for hierarchical federated learning,'' \emph{IEEE Transactions on
  Parallel and Distributed Systems}, vol.~33, no.~3, pp. 536--550, 2021.

\bibitem{hui2022quality_133}
D.~Hui, L.~Zhuo, and C.~Xin, ``Quality-aware incentive mechanism design based
  on matching game for hierarchical federated learning,'' in \emph{IEEE INFOCOM
  2022-IEEE Conference on Computer Communications Workshops (INFOCOM
  WKSHPS)}.\hskip 1em plus 0.5em minus 0.4em\relax IEEE, 2022, pp. 1--6.

\bibitem{9562751}
Y.~Jin, L.~Jiao, Z.~Qian, S.~Zhang, and S.~Lu, ``Budget-aware online control of
  edge federated learning on streaming data with stochastic inputs,''
  \emph{IEEE Journal on Selected Areas in Communications}, vol.~39, no.~12, pp.
  3704--3722, 2021.

\bibitem{Boxin}
\BIBentryALTinterwordspacing
B.~Zhao, L.~Wang, Z.~Liu, Z.~Zhang, J.~Zhou, C.~Chen, and M.~Kolar, ``Adaptive
  client sampling in federated learning via online learning with bandit
  feedback,'' \emph{Journal of Machine Learning Research}, vol.~26, no.~8, pp.
  1--67, 2025. [Online]. Available:
  \url{http://jmlr.org/papers/v26/24-0385.html}
\BIBentrySTDinterwordspacing

\bibitem{shukla2021federated_221}
S.~Shukla and N.~Srivastava, ``Federated matched averaging with
  information-gain based parameter sampling,'' in \emph{Proceedings of the
  First International Conference on AI-ML Systems}, 2021, pp. 1--7.

\bibitem{9593194}
C.-H. Hu, Z.~Chen, and E.~G. Larsson, ``Device scheduling and update
  aggregation policies for asynchronous federated learning,'' in \emph{2021
  IEEE 22nd International Workshop on Signal Processing Advances in Wireless
  Communications (SPAWC)}, 2021, pp. 281--285.

\bibitem{liu2024checkpoint}
D.~Liu, Z.~Wang, B.~Wang, W.~Chen, C.~Li, Z.~Tu, D.~Chu, B.~Li, and D.~Sui,
  ``Checkpoint merging via bayesian optimization in llm pretraining,''
  \emph{arXiv preprint arXiv:2403.19390}, 2024.

\bibitem{he2020group}
C.~He, M.~Annavaram, and S.~Avestimehr, ``Group knowledge transfer: Federated
  learning of large cnns at the edge,'' \emph{Advances in neural information
  processing systems}, vol.~33, pp. 14\,068--14\,080, 2020.

\bibitem{Howard2017MobileNetsEC}
\BIBentryALTinterwordspacing
A.~G. Howard, M.~Zhu, B.~Chen, D.~Kalenichenko, W.~Wang, T.~Weyand,
  M.~Andreetto, and H.~Adam, ``Mobilenets: Efficient convolutional neural
  networks for mobile vision applications,'' \emph{ArXiv}, vol. abs/1704.04861,
  2017. [Online]. Available:
  \url{https://api.semanticscholar.org/CorpusID:12670695}
\BIBentrySTDinterwordspacing

\bibitem{Hoyer}
P.~O. Hoyer, ``Non-negative matrix factorization with sparseness constraints,''
  \emph{J. Mach. Learn. Res.}, vol.~5, p. 1457–1469, 12 2004.

\bibitem{Krizhevsky2009LearningML}
\BIBentryALTinterwordspacing
A.~Krizhevsky, ``Learning multiple layers of features from tiny images,'' 2009.
  [Online]. Available: \url{https://api.semanticscholar.org/CorpusID:18268744}
\BIBentrySTDinterwordspacing

\end{thebibliography}
\end{document}